\documentclass[conference]{IEEEtran}
\IEEEoverridecommandlockouts
\usepackage{cite}
\usepackage{amsmath,amssymb,amsfonts}
\usepackage{algorithmic}
\usepackage{graphicx}
\usepackage{textcomp}
\usepackage{xcolor}
\usepackage{array}
\usepackage{hyperref}
\usepackage{multirow}

\newcolumntype{P}[1]{>{\centering\arraybackslash}p{#1}}
\newcolumntype{M}[1]{>{\centering\arraybackslash}m{#1}}

\def\BibTeX{{\rm B\kern-.05em{\sc i\kern-.025em b}\kern-.08em
    T\kern-.1667em\lower.7ex\hbox{E}\kern-.125emX}}
    
\usepackage{balance}
\usepackage{lastpage}

\makeatletter
    \def\balanceissued{unbalanced}
    \let\oldbibitem\bibitem
    \def\bibitem{%
        \ifnum\thepage=\lastpage@lastpage%
            \expandafter\ifx\expandafter\relax\balanceissued\relax\else%
                \balance%
                \gdef\balanceissued{\relax}\fi%
            \else\fi%
        \oldbibitem}
\makeatother

\begin{document}

\title{Neural Grammatical Error Correction for Romanian\\
}

\author{\IEEEauthorblockN{Teodor-Mihai Cotet}
\IEEEauthorblockA{\textit{Computer Science Department}\\
\textit{University Politehnica of Bucharest}\\
Bucharest, Romania \\
teodor\_mihai.cotet@stud.acs.upb.ro}
\and
\IEEEauthorblockN{Stefan Ruseti}
\IEEEauthorblockA{\textit{Computer Science Department} \\
\textit{University Politehnica of Bucharest}\\
Bucharest, Romania \\
stefan.ruseti@upb.ro}
\and
\IEEEauthorblockN{Mihai Dascalu}
\IEEEauthorblockA{\textit{Computer Science Department} \\
\textit{University Politehnica of Bucharest}\\
Bucharest, Romania \\
mihai.dascalu@upb.ro}
}

\maketitle
\begin{abstract}
Resources for Grammatical Error Correction (GEC) in non-English languages are scarce, while available spellcheckers in these languages are mostly limited to simple corrections and rules. In this paper we introduce a first GEC corpus for Romanian consisting of 10k pairs of sentences. In addition, the German version of ERRANT (ERRor ANnotation Toolkit) scorer was adapted for Romanian to analyze this corpus and extract edits needed for evaluation. Multiple neural models were experimented, together with pretraining strategies, which proved effective for GEC in low-resource  settings. Our baseline consists of a small Transformer model trained only on the GEC dataset ($F_{0.5}$ = 44.38), whereas the best performing model is produced by pretraining a larger Transformer model on artificially generated data, followed by finetuning on the actual corpus ($F_{0.5}$ = 53.76). The proposed method for generating additional training examples is easily extensible and can be applied to any language, as it requires only a POS tagger.
\end{abstract}

\begin{IEEEkeywords}
Grammatical Error Correction, Transformer, Romanian, low-resource language, ERRANT
\end{IEEEkeywords}

\section{Introduction}
\label{intro}

Grammatical Error Correction (GEC) considers correcting all types of language mistakes from a given text. These errors are not limited to  grammatical mistakes, but also include punctuation, spelling, and word choice errors. For simplicity, datasets for GEC are generally built at the sentence level \cite{ng2014conll}, meaning that a dataset should provide pairs of original/corrected sentences. Optionally, datasets also provide multiple corrected sentences for the same original one. For example, the JFLEG (JHU FLuency-Extended GUG)~\cite{napoles2017jfleg} corpus provides four references for each original sentence to enable  a better evaluation, while ensuring more freedom in the corrections so that the generated texts sound more natural.

GEC systems can be evaluated by comparing edits from the gold corpus with the edits extracted from the predicted sentence. An edit is defined as the minimal set of changed words required to correct one mistake. It is represented as a pair: the word(s) from the original sentence, and the word(s) from the corrected sentence. The extraction of edits can be performed automatically by using the Damerau-Levenshtein distance, together with merges for multi-token edits~\cite{felice2016automatic}. The substitution cost in Damerau-Levenshtein distance is based heavily on linguistic features in order for the substitution of similar words to have priority compared to insertions and deletions. The edits are then extracted by following the optimal path in the cost matrix. On top of that, the ERRANT (ERRor ANnotation Toolkit) ~\cite{bryant2017automatic} system classify the extracted edits using a rule-based system based on the POS tags. ERRANT is also also our selected choice for extracting and classifying edits. Note that there is no unique way of extracting these edits. Depending on how one defines the costs in the Damerau-Levenshtein operations one can obtain different edits. Generally, the automatic extraction of these edits is performed such that it is similar to the edits marked by human experts~\cite{felice2016automatic}.

Most progress in Grammatical Error Correction focused on English datasets. There is a limited number of other languages for which GEC systems were developed, namely: Arabic~\cite{zaghouani2014large}, German~\cite{boyd2018using}, Russian~\cite{rozovskaya2019grammar}, Czech~\cite{naplava2017natural}, Chinese~\cite{yu2014overview} and Japanese \cite{mizumoto2011mining}. A review of these datasets is presented by N{\'a}plava and Straka~ \cite{naplava2019grammatical}, together with state-of-the-art results for Czech, Russian, and German languages.

Our aim is to introduce a GEC corpus for Romanian consisting of approximately 10k pairs of sentences (further referred to as gold corpus), as well as a neural baseline model, together with improvements from an artificially generated dataset. Moreover, the ERRANT toolkit was adapted for Romanian in order to evaluate the models. The corpus contains sentences or phrases from Romanian TV or radio shows, corrected by the National Audiovisual Council (NAC\footnote{http://www.cna.ro/}) which supervises these entities, as well as some additional sentences in order to better mimic the written mistakes. This corpus is further divided into three groups, capturing the differences in the types of sentences and mistakes. Our baseline consists of a small Transformer model~\cite{vaswani2017attention} trained only on the gold corpus. We also experimented with a Transformer-base model pretrained on artificially generated data from Wikipedia. Our approach is mainly inspired from the work of Grundkiewicz et al.~\cite{grundkiewicz2019neural} whose model ranked first in the BEA 2019 Shared Task on GEC~\cite{bryant2019bea}, low-resources task. In this track, participants could only use the development pairs of sentences (precisely 4,384) from the Cambridge English Write \& Improve (W\&I) and LOCNESS corpus (W\&I+LOCNESS), which is comparable in size to our corpus. After pretraining, two methods of finetuning the model on the gold corpus were tried, together with three strategies for decoding, further on presented in detail.

We also release a 5-gram language model (used for decoding) in the ARPA\footnote{https://cmusphinx.github.io/wiki/arpa/} format trained on 30 million sentences filtered from Wikipedia, together with the 10 million pairs of sentences used for pretraining. Moreover, the simplified German version of ERRANT~\cite{bryant2017automatic} was adapted for Romanian. All the resources and models can be found in the Github repository\footnote{https://github.com/TeodorMihai/RoGEC} and are available under an Apache 2 license.

The paper is structured as follows. The second section presents state-of-the-art methods and corpora for GEC, while the third section introduces our gold corpus. The fourth section presents our method, with corresponding evaluation metrics, followed by results using various Transformer-based architectures. The last section consists of conclusions and envisioned future work.

\section{Related Work}
Similar to machine translation, state-of-the-art systems for GEC shifted from phrase-based~\cite{junczys2016phrase} approaches to pure neural methods~\cite{xie2016neural,junczys2018approaching}. The best architectures for the 2014 benchmark on GEC~\cite{ng2014conll} are all Transformer-based~\cite{zhao2019improving,kiyono2019empirical}. The most recent approaches~\cite{kaneko2020encoder} incorporate pretrained masked language models, such as BERT~\cite{devlin2018bert}. Integrating BERT into GEC task proves difficult because the hidden representation of the incorrect words is very similar to the representation of correct words. Kaneko et al.~\cite{kaneko2020encoder} overcome this problem by finetuning the pretrained BERT on the GED task (grammatical error detection), and then use BERT as additional features to the model. The 2014-CoNLL shared task on GEC allows the use of larger corpora available for English, but not for other languages. Thus, methods which use only limited datasets are further presented. 

The low-resource scenario in GEC is generally defined by the rules of the BEA 2019 Shared Task on GEC~\cite{bryant2019bea} on the respective track. These rules prohibit the use of specially annotated GEC datasets, other than the official dataset. However, the rules do not prohibit the use of alternative ways of generating or obtaining training data. Therefore, artificially generated data or filtered Wikipedia edits are allowed. 

There are several options to handle low-resource scenarios in GEC. Bryant and Briscoe~\cite{bryant2018language} present an approach that only needs a spellchecker, a strong language model, and a minimal dataset of annotated entries ($\sim$1000 sentences). Their algorithm iteratively improves the normalized log probability given by a language model for the sentence. At each iteration, each word is considered for substitution by choosing one single substitution that maximizes the log probability until a threshold is reached. The substitutions are permitted only from the confusion set of the word, with additional words for articles and prepositions. The confusion set represents the suggestions made by a spellchecker for a given word. This method proves the power of language models that achieve competitive results, with no additional corpora.

A method of constructing additional training corpora is to extract parallel sentences from Wikipedia revision history, filtering them in numerous ways, such that that only grammatical revisions are selected~\cite{lichtarge2019corpora}. This method was successfully applied for German, which has only a limited dataset for GEC~\cite{boyd2018using}. In this method, parallel sentences from Wikipedia were filtered with the help of ERRANT which produced edits and their respective categories. Filtered Wikipedia edits proved beneficial in the training process, when compared to using all unfiltered edits.

Another method of augmenting the training set is to modify a monolingual corpus by introducing synthetic errors; these sentences are generally used for pretraining. Yuan and Felice~\cite{yuan2013constrained} introduced errors by extracting edits from the main corpus and substituting them in the monolingual corpus. Grundkiewicz et al.~\cite{grundkiewicz2019neural} used the substitution of a word with a member of its corresponding confusion set generated by a spellchecker as their main method of error generation.

\section{Gold Corpus}
We introduce, to the best of our knowledge, the first corpus for Romanian GEC. The Romanian National Audiovisual Council Corpus (\textit{RONACC}) is a native (first language) corpus considering corrected sentences spoken on Romanian TV and radio shows, together with additional sentence pairs that represent more common written mistakes. 

The corpus is partitioned into 3 groups: \textit{NAC-phrases}, \textit{NAC-sentences}, and \textit{W-sentences}. The first two categories are sentences from the spoken language.  NAC-phrases contains phrases that do not include a verb, while NAC-sentences are well formed sentences. The last group, W-sentences, consists of well formed sentences with written mistakes. All phrases in the corpus contain enough information to be corrected without any additional context. 

The corpus was split into train (70\%), dev (15\%), and test (15\%) partitions. Each set contains an equal amount of examples from each aforementioned group. Statistics provided by ERRANT on the corpus are presented in Table~\ref{tab:corpus-stats}, together with details on the artificially generated corpus presented later on. 

\begin{table*}[ht]
\centering
\caption{RONACC statistics.}
\begin{tabular}{ |l|r|r|r|r|r|r|r| }
\hline
      & \textbf{Sentences} &  \textbf{POS(14)} & \textbf{MORPH} & \textbf{ORTH} & \textbf{SPELL} & \textbf{ORDER} & \textbf{OTHER}\\
 \hline
    NAC-phrases& 2,639  & 3,741 & 184 & 127 & 780 & 6 & 775\\
    \hline
    NAC-sentences& 4,475 & 7,076 & 172 & 108 & 797 & 90 & 1,438\\
    \hline
    W-sentences& 3,005 & 4,921 & 189 & 758 & 1948 & 0 & 1,061\\
    \hline
    \textit{Total gold corpus} & \textit{10,119} & \textit{15,738 (65.1\%)} & \textit{545 (2.2\%)} & \textit{993 (4.1\%)} & \textit{3,525 (14.5\%)} & \textit{96 (0.3\%)} & \textit{3,274 (13.5\%)} \\
    \hline
    \hline
    Artificial corpus (subsample) & 300k (out of 10M) & 53.9\% & 1.8\% & 1.2\% & 12.1\% & 2.7\% & 27.9\% \\ 
\hline
\end{tabular}
\label{tab:corpus-stats}
\end{table*}

\subsection{ERRANT}
An automated classifications of  errors into several types was performed  to better understand the corpus, based on a simplified ERRANT version for German~\cite{boyd2018using}. Our adapted version uses only UD POS tags, instead of language-specific tags. Our modifications in terms of rules and types consider the elimination of the German-specific STTS (Stuttgart–Tubingen Tagset) tag \textit{TRUNC}, as well as the the elimination of \textit{CONTR} category, standing for contraction errors. The dictionary from the Hunspell spellchecker\footnote{https://github.com/titoBouzout/Dictionaries} was used as the word list for ERRANT to detect spelling errors. POS tagging was performed with a Romanian SpaCy\footnote{https://spacy.io} model integrated in the ReaderBench~\cite{gutu2016readerbench} framework. Similar to the German version, the POS category includes 14 UD POS tags, with an extra \textit{FORM} tag (e.g., \textit{NOUN:FORM}) for tokens with the same lemma. The \textit{MORPH} category refers to errors related to word form, \textit{ORTH} to capitalization and whitespace errors, \textit{SPELL} is a spelling error, highlighting words not found in the dictionary with more than half overlapping characters with their corrected versions, whereas \textit{ORDER} refers to errors related to the reordering of tokens. Examples from the gold corpus classified by ERRANT type can be observed in Table~\ref{tab:examples-corpus}.

\begin{table}[t]
    \centering
    \caption{Examples from the gold corpus.}
    \begin{tabular}{|p{2.1cm}|p{5.5cm}|}
    \hline
    \textbf{ERRANT type} & \textbf{Example} \\
        \hline
        MORPH & în cazul unei paciente [\textbf{internată} $\rightarrow$ internate] joi \\
        \hline
        ORTH & Acum, în [\textbf{camera deputaților} $\rightarrow$ Camera Deputaților]\\
        \hline 
        SPELL & treceți la ceea ce [\textbf{vroiați} $\rightarrow$ voiați] să ziceți \\
        \hline 
        ORDER & Nu [\textbf{mai o} $\rightarrow$ o mai] da cotită  \\
        \hline 
        OTHER & permis de [\textbf{port-armă} $\rightarrow$ portarmă] \\
        \hline
        POS:NOUN &  O rotiță care să-și aducă [\textbf{aportul} $\rightarrow$ contribuția].  \\
        \hline 
        POS:NOUN:FORM & să vedem cum a fost din [\textbf{punct} $\rightarrow$ punctul] de vedere al organizării [...] \\
        \hline 
        POS:VERB & opoziția ar putea [\textbf{demara} $\rightarrow$ începe] procedura de suspendare \\
        \hline 
        POS:VERB:FORM & Omul negru poate [\textbf{fii} $\rightarrow$ fi] folosit metaforic [...] \\
        \hline
        POS:ADJ & Mulți se tem astăzi să se întâlnească cu personaje dubioase sau într-un context [\textbf{aiurea} $\rightarrow$ nepotrivit].\\
        \hline
        POS:ADJ:FORM & E un pic [\textbf{ambiguu} $\rightarrow$ ambiguă] definirea termenului mită aici \\
        \hline
        POS:ADV & Și te costă și foarte, foarte [\textbf{ieftin} $\rightarrow$ puțin]\\
        \hline 
        POS:PRON & o să trebuiască să dați niște mesaje, [\textbf{\_} $\rightarrow$ niște] telefoane [...] \\
        \hline
        POS:PRON:FORM & Pe [\textbf{aceiași} $\rightarrow$ aceeași] bandă circulai? \\
        \hline
        POS:DET & șeful [\textbf{\_} $\rightarrow$ unui] aerodrom privat din Comana \\
        \hline 
        POS:DET:FORM & de liderii PSD, PNL și [\textbf{a} $\rightarrow$ ai] minorităților naționale \\
        \hline
        POS:ADP & 80\% [\textbf{din} $\rightarrow$ dintre] victimele traficanților \\
        \hline 
        POS:CCONJ & dânsa ca [\textbf{și} $\rightarrow$ \_] persoană luptătoare \\
        \hline
        POS:PUNCT & lovește mingea înapoi[\textbf{\_} $\rightarrow$ ,]dar au început să se apropie \\
        \hline 
    \end{tabular}
    \label{tab:examples-corpus}
\end{table}

\subsection{Artificially Generated Data}
All individual sentences from the Romanian Wikipedia were selected to generate synthetic errors. Unfortunately, Wikipedia dumps are noisy, including many sentences that have links or are not well formed (no clear beginning or ending) or are not even sentences, but include just a few links or metadata. The following filtering criteria were used, resulting in  10 million sentences:

\begin{itemize}
    \item Sentence starts with an uppercase letter; this rule filters not well-written sentences.
    \item Quotes signs and no sub-string indicating links are present in the sentence (e.g. "www."); this rule has the effect of filtering non-sentences.
    \item The parenthesis beginning and ending for the two types (i.e., "(" and "[") are in equal number; this rule just filters well-written sentences.
	\item The last punctuation sign of the sentence is an end mark (i.e., "?!.") and it is not coming from an abbreviation; this rule selects complete sentences. 
	\item The ratio of diacritics characters to non-diacritics characters is above 1\%; many digital texts in Romanian do not have diacritics, including texts from Wikipedia.
	\item The ratio between ASCII and diacritics characters and other types of characters is not larger than 2.5\%; this rule is designed to filter out metadata. 
	\item The sentence contains more than 8 words; the main goal of this filter is to make sure that sentences contain enough context for a model to be able to correct a misspelled word. 
\end{itemize}
	
Afterwards, each sentence was automatically altered to simulate possible errors. The following methodology of altering sentences was used, inspired by Grundkiewicz et al.~\cite{grundkiewicz2019neural}:
\begin{enumerate}
    \item Sample $p_{err}$ from a normal with mean of 0.15 and standard deviation of 0.2.
    \item Multiply $p_{err}$ by the length of the sentence and round it, getting $N_{changed}$.
	\item Choose the words to change by sampling uniformly $N_{changed}$ times without replacement from the words of the sentence.
	\item For each sampled word $w_i$ choose the following:
	\begin{enumerate}
	    \item Substitute $w_i$ with a random word from its confusion set (limited to the top 20 suggestions) generated by a spellchecker with probability of 0.7.
	    \item Delete $w_i$ with probability 0.1.
	    \item Insert a random word after $w_i$ with probability of 0.1.
	    \item Swap $w_i$ with an adjacent word with probability of 0.1.
	\end{enumerate}
	\item Apply the same operations at 4 for characters, changing 10\% of the words in the sentence with the same probabilities for operations.
\end{enumerate}

The errors introduced at step 5 have the main goal of making the model capable of correcting spelling errors. The confusion set is used so that the correction task remains a difficult one, that will need at least a partial understanding of the context; Aspell \footnote{http://aspell.net/} was used for spellchecking. The value of $p_{err}$ was selected to simulate the proportion of mistakes from GEC datasets. 

Statistics about the artificially generated corpus are presented in Table~\ref{tab:corpus-stats}. Only a subset of 300 thousand pairs out of the 10 million sentences were analyzed, but we expect this numbers to reflect the general composition of the artificially generated corpus, given that the same rules of alteration were applied uniformly. The relative similarity in error distributions of the two corpora is another possible explanation of the effectiveness of this method of pretraining. GEC training on a corpus other than the gold corpus is sensible to the error rates differences in the two corpora. Adjusting the overall error rate between these two datasets (by eliminating some sentences from the augmentation corpus) proved beneficial for English \cite{junczys2018approaching}. However, detailed ERRANT edits types for POS category are not as similar, as can be noticed in Table \ref{tab:errant-pos}. Notably, the (:\textit{FORM}) types have a significant lower contribution in the artificial corpus compared the the gold one. This will encourage models not to be conservative in corrections, modifying word forms. Moreover, the \textit{NOUN} category is disproportionately larger, while the \textit{PUNCT} category is considerably lower; nevertheless, the later changes are simpler to correct and not that many examples are required.

\begin{table}[t]
    \centering
    \caption{Detailed ERRANT edits types for POS category.}
    \begin{tabular}{|c|c|c|}
        \hline
        \textbf{POS} & \textbf{Gold corpus (\%)} & \textbf{Artificial Corpus (\%)} \\
        \hline
        NOUN & 5.34\% &  45.64\% \\
        \hline
        NOUN:FORM & 14.55\% & 4.26\% \\
        \hline
        VERB & 7.25\% & 14.17\% \\
        \hline
        VERB:FORM & 5.99\% & 3.08\% \\
        \hline
        ADJ & 1.23\% & 8.17\% \\
        \hline
        ADJ:FORM & 1.60\% & 1.34\% \\
        \hline
        ADV & 2.92\% &  2.59\% \\
        \hline
        ADV:FORM & 0.00\% & 0.02\% \\
        \hline 
        PRON & 6.91\% & 5.39\% \\
        \hline 
        PRON:FORM & 0.75\% & 0.17\% \\
        \hline 
        DET & 9.13\% & 2.90\% \\
        \hline 
        DET:FORM &  1.05\% & 0.13\% \\
        \hline 
        ADP & 15.98\% & 9.55\% \\
        \hline 
        CCONJ & 2.20\% & 2.16\% \\
        \hline 
        PUNCT & 23.58\% & 0.36\% \\
        \hline
    \end{tabular}
    \label{tab:errant-pos}
\end{table}

\section{Method}
Our proposed architectures are based on the Transformer model~\cite{vaswani2017attention}, a general purpose architecture based mostly on self-attention blocks between different units (i.e., tokens for subwords in NLP). Transformer is a sequence-to-sequence model, meaning it contains an encoder and a decoder, which achieved state-of-the-art results on multiple tasks (e.g., machine translation, named entity recognition) at the time of its publication. Both encoder and decoder construct an internal representation of the sentence and of words by using attention mechanisms in relation to the other words from the sentence. 

\subsection{Model}
A small Transformer model trained only on the gold corpus was used as baseline (approx. 1.5 mil parameters); this model will be referred to as "Transformer-tiny". A larger model ("Transformer-base") was also evaluated - this model was first pretrained on artificially altered sentences, before using the gold corpus for further training. Open vocabulary issues were addressed using a word-piece vocabulary \cite{wu2016google}. There is no pair of embeddings that are shared among different layers. The default architecture of the Transformer-base model was used~\cite{vaswani2017attention}, preserving all the hyperparameters and dimensions, as well the learning rate scheduling. The dictionary consists of 32,768 word-pieces constructed from the entire artificially generated corpus. The training was performed until the loss on the development set was not increasing substantially. The development set consisted of 3\% of the artificially generated corpus.

A comparison between the parameters used by the two models can be seen in Table \ref{tab:parameters}. The models may not seem that different in terms of size, given each parameter individually. However, Transformer-tiny consists overall of approx. 15k parameters, while Transformer-base has around 94 million parameters. This difference is mostly due to the massive decrease in the vocabulary size. The Adam optimizer was used in both architectures, with a warm-up schedule of the learning rate.

\begin{table}[t]
    \centering
    \caption{Parameter comparison between Transformer-tiny and Transformer-base.}
    \begin{tabular}{|c|c|c|}
        \hline
        \textbf{Parameters} & \textbf{Transformer-tiny} & \textbf{Transformer-base} \\
        \hline
        Encoder layers & 3 & 6 \\
        \hline
        Decoder layers & 3 & 6 \\
        \hline
        Embeddings size & 128 & 512 \\
        \hline
        Filter size & 128 & 2048 \\
        \hline
        Attention heads & 2 & 8 \\
        \hline
        Dictionary size & 2k & 32k \\
        \hline
        Dropout & 0.2 & 0.1 \\
        \hline
    \end{tabular}
    \label{tab:parameters}
\end{table}

Two strategies to continue training only on the gold corpus were experimented: \textit{retraining} and \textit{finetuning}. Retraining resets all parameters of the optimizer and its learning rate before training, while finetuning continues training with the preserved parameters. However, choosing one of these strategies is not straightforward beforehand. The retraining procedures might result in an increase the the model's capacity to forget its pretraining on artificial data, thus increasing the risk of overfitting. The finetuning approach might not be able to change the model enough to adapt on the gold corpus. Both procedures finetune all layers in the model, with no parameters being frozen.

Decoding was performed using a beam search with a beam of size 8 in all experiments. This beam size was chosen as a compromise between accuracy and speed, a value in a common range used in similar tasks for decoding  ~\cite{naplava2017natural,grundkiewicz2019neural}. A Kneser-Ney smoothed 5-gram language model was used for re-ranking of the beams. The language model was build using the KenLM toolkit~\cite{heafield2013scalable}. All experiments use a normalized probability given by the language model. The weight of the normalized probability of the language model was the same as the weight of the beam search probabilities and it was not grid-searched. Moreover, we experimented with a length-normalized beam search.

\subsection{Evaluation}
Metrics frequently used in machine translation, for example the BLEU score~\cite{papineni2002bleu}, do not provide good indications of the model's performance due to the high amount of identical words between the predicted sentence and the gold sentence in which words are simply copied. Thus, metrics which only take into account the edits are frequently employed in GEC. The ERRANT system~\cite{bryant2017automatic} performs the extraction and scoring of edits for the GEC task. First, the system extracts edits using a rule-based system which is mainly based on the UD POS tags, because most errors are based on POS tags. The system consists of 50 such rules. 

Once the edits are extracted, the evaluation becomes easier. ERRANT uses a simple process of comparing the edits between those of the original and gold sentences, and those of original and predicted sentences. If an edit of the latter is found in the former, then it is considered a true positive. False-positives and false-negatives follow a similar process. Afterwards, precision and recall are computed, resulting in the combined $F_{0.5}$ score in which the precision is 2 times more important than recall. This emphasize on precision is due to the requirements of GEC systems that should suggest corrections only when they are quite sure about it. The omission of a correction, due to a low recall, is not as important.

\section{Results}
Table \ref{tab:model-results} introduces results for various configurations. The Transformer-tiny model was only trained on the RONACC corpus, while the larger ones also used the artificially generated dataset. The baseline result proves the power of the Transformer model, even when using a relative small number of parameters. The model for the baseline was trained only a few hours on a modest GPU (i.e., NVidia GTX 1050). At the same time, we experimented with a dropout rate of 0.3, but no improvements were noticed. Moreover, the baseline presents a comparable precision to the finetuned model, and an $F_{0.5}$ score similar to the retrained model.

\begin{table*}[t]
\begin{center}
\caption{Results on RONACC corpus.}
\begin{tabular}{ |c|l|c|c|c|}
    \hline
      \textbf{Model} & \textbf{Decoding} & \textbf{P} & \textbf{R} &  {$\boldsymbol F_{0.5}$}\\
 \hline
    Transformer-tiny & beam search & 53.53 & 26.36 & 44.38\\
 \hline 
 \multirow{3}{4cm}{\centering Transformer-base artificial data only} 
                        & beam search     & 16.74 & 16.18 & 16.62\\
                        \cline{2-5}
                        & beam search + LM & 17.33 & 17.27 & 17.32\\
                        \cline{2-5}
                        & beam search normalized + LM & 17.30 & 17.27 & 17.30\\
\hline
  \multirow{3}{4cm}{\centering Transformer-base Retraining}     
                        & beam search & 45.47 & 44.84 & 45.35\\
                        \cline{2-5}
                        & beam search + LM & 41.89 & 43.71 & 42.24\\
                        \cline{2-5}
                        & beam search normalized + LM & 41.95 & 43.80 & 42.31\\
 \hline                               
   \multirow{3}{4cm}{\centering Transformer-base Finetuning}            
                        & beam search &  \textbf{56.05} & \textbf{46.19} & \textbf{53.76}\\
                        \cline{2-5}
                        & beam search + LM & 50.68 & 45.39 & 49.52\\
                        \cline{2-5}
                        & beam search normalized + LM & 51.06 & 45.43 & 49.83\\
\hline
\end{tabular}
\label{tab:model-results}
\end{center}
\end{table*}

Similar to the findings of Grundkiewicz et al.~\cite{grundkiewicz2019neural}, the finetuning technique proved superior to retraining, with an $F_{0.5}$ score improvement of 8 points. The difference is mainly due to a higher precision. Re-ranking using the language model notably decreased the overall performance, with the exception of the model trained only on artificially generated sentences. This is due to two main reasons. The first one is the small size of the training dataset for the language model, only 30 million sentences. The second one is the pre-defined weight value of the language model in the re-ranking formula, compared to the weight of the beam search probability. Normally, this weight is grid-searched on the development set, but we did not posses the required hardware resources to perform the grid-search. Thus, the weights for both the language model and the beam search probability were fixed to 1. Normalization of the beam search did not affect the scores in a significant way.

Results of the baseline and finetuned models for the three RONACC groups are presented in Table \ref{tab:group-stats}. The highest difference of $F_{0.5}$ scores between the baseline and the finetuned model is visible on the W-sentences group. We expect this difference is due to the higher similarity between the W-sentences group, which mimics written mistakes more, and the synthetic sentences. The W-sentences group contains sentences from Wikipedia, thus the language used is very similar to the language in the artificially generated sentences.

\begin{table*}[ht]
\begin{center}
\caption{Results on RONACC groups.}
\begin{tabular}{|c|c|c|c|c|c|c|c|c|c|}
    \hline
    \textbf{Model} &\multicolumn{3}{c|}{\textbf{NAC-phrases}} & \multicolumn{3}{c|}{\textbf{NAC-sentences}} &\multicolumn{3}{c|}{\textbf{W-sentence}}\\
    \cline{2-10}
                                & \textit{P} & \textit{R} & $F_{0.5}$ & \textit{P} & \textit{R} & $F_{0.5}$ & \textit{P} & \textit{R} & $F_{0.5}$\\
    \hline 
    Transformer-tiny                    & 51.3 & 28.1 & 44.1 & 52.7 & 21.9 & 41.2 &55.3 & 29.5 & 47.1\\
    \hline
    Transformer-base with Finetuning & 55.1 & 41.2 & 51.6 & 46.5 & 32.0 & 42.6 &62.7 & 62.6 & 62.7\\
\hline
\end{tabular}
\label{tab:group-stats}
\end{center}
\end{table*}

Examples of predictions are presented in Table \ref{tab:examples-predictions}. Spelling errors are generally well handled (e.g., "ultimile" - eng. "last" - is corrected to "ultimele"), as well as the Romanian-specific problem of multiple ‘i’ (e.g., "oameni" - eng. "people" is corrected to "oamenii"), which depends on the plural and if the noun is articulated or not. Word order is sometimes not corrected (e.g., "mai ne" versus "ne mai").

The hyphen sign is used in Romanian to separate to syntax words merged into one. This is sometimes a difficult task as context defines whether a hyphen is required or not. Some of the uses are however very easy to figure out. The model generally can put the hyphen correctly, or eliminate it if that is the case (e.g., "să-l" vs "săl").

The NAC-sentences group benefited least from pretraining on the artificially generated dataset and has the lowest $F_{0.5}$ for both baseline and finetuned models. This is due to the nature of the mistakes and language from this group compared to the artificially generated corpus. This group includes changing some nouns altogether because their use is not idiomatic, or is not formal enough. Correcting non-idiomatic mistakes remains an open problem for small corpora GEC. In addition, English words were also encountered. For example, the word "tricks" appears in the corpus in the English form. Instead of using the proper Romanian noun for the word "tricks", the model changes it to a similar Romanian noun in terms of characters overlapping. Given the popularity of  English words in non-English languages, we suspect that mixing  English words into the training corpus would benefit the system.

Another problem is the informal nature of the spoken language, compared to the more formal sentences from the synthetic corpus. For instance, the informal, but correct, Romanian word for "a voma" (eng. "to puke") is changed by the system to different verb - "a vota" (eng. "to vote"). These errors affect both NAC-phrases and NAC-sentences groups. This problem could probably be tackled by inserting more informal, but correctly written, sentences in the training set.

\begin{table*}[ht]
    \centering
    \caption{Examples of predictions.}
    \begin{tabular}{|c|p{4cm}|p{4cm}|p{4cm}|}
    \hline
    \textbf{Group} & \textbf{Original} & \textbf{Gold} & \textbf{Predicted} \\
        \hline
        \multirow{3}{2cm}[-3.5em]{\centering W-sentences} &
        \textbf{Oameni} nu îi judecă pe
\textbf{bărbații} că nu sunt tați
buni & \textbf{Oamenii} nu îi judecă pe
\textbf{bărbați} că nu sunt tați buni & \textbf{Oamenii} nu îi judecă pe
\textbf{bărbații} că nu sunt tați buni \\
        \cline{2-4}
                & Terminând cu aventurile
sale, Ben \textbf{șia} scos
omnitrix-ul și a crescut
de la un băiat mic \textbf{întrun}
adolescent pe care te
poți baza. & 
Terminând cu aventurile
sale, Ben \textbf{și-a} scos omnitrix-
ul și a crescut de la un băiat
mic \textbf{într-un} adolescent pe
care te poți baza. & Terminând cu aventurile sale,
Ben \textbf{și-a} scos omnitrix-ul și a
crescut de la un băiat mic \textbf{într-
un} adolescent pe care te poți
baza. \\
\cline{2-4}
& Mohamed care a sosit la
Londra luna trecută
pentru operație\textbf{ v-a fii}
externat mai târziu din
spital.
& Mohamed, care a sosit la
Londra luna trecută pentru
operație, \textbf{va fi} externat mai
târziu din spital.
& Mohamed care a sosit la
Londra luna trecută pentru
operație, \textbf{va fi} externat mai
târziu din spital.\\
\hline 
\multirow{3}{2cm}{\centering NAC-phrases} 
& în \textbf{ultimile} trei zile & în \textbf{ultimele} trei zile & în \textbf{ultimele} trei zile \\
\cline{2-4}
& Pe terenurile \textbf{Grădinei}
Zoologice & Pe terenurile \textbf{Grădinii}
Zoologice & Pe terenurile \textbf{Grădinii}
Zoologice \\
\cline{2-4}
& \textbf{una}-două grade Celsius & \textbf{unu}-două grade Celsius & \textbf{una}-două grade Celsius \\
\hline
\multirow{3}{2cm}[-2.2em]{\centering NAC-sentences}
& ma vor chema \textbf{pt} a 41-a
oara & mă vor chema \textbf{pentru} a 41-a
oară & mă vor chema \textbf{pentru} a 41-a
oară\\ 
\cline{2-4}
& Busuioc acuzat că s-a îmbătat și \textbf{vomat} într-un taxi & 
Busuioc, acuzat că s-a îmbătat și \textbf{a vomat} într-un taxi &
Busuioc, acuzat că s-a îmbătat și \textbf{votat} într-un taxi\\
\cline{2-4}
& a luat deja o decizie și să
sperăm că va fi \textbf{de
comun acord} cu
intențiile voastre & 
a luat deja o decizie și să
sperăm că va fi \textbf{în acord} cu
intențiile voastre
& a luat deja o decizie și să
sperăm că va fi \textbf{de acord} cu
intențiile voastre \\
\cline{2-4}
& există totuși niște \textbf{trickuri} & există totuși niște \textbf{secrete} & există totuși niște \textbf{tricouri} \\
\hline
    \end{tabular}
    \label{tab:examples-predictions}
\end{table*}

\section{Conclusions and Future Work}
We provide a neural baseline on the RONACC corpus for Romanian GEC. We also improved upon this baseline by using tested techniques for low-resource English GEC~\cite{grundkiewicz2019neural}. An artificially generated corpus was used to pretrain a large Transformer model, experimenting with 2 techniques for further training on the gold corpus: finetuning and retraining. Moreover, the ERRANT system may now be used to evaluate systems on Romanian GEC or to infer statistics about any GEC datasets.

We also experimented with the integration of a pretrained BERT model as an encoder; however, due to its poor performance, we did not include it in the final experiments. The usual methods of integrating BERT into NLP tasks do not perform well for GEC. The hidden representations for incorrect words become very similar to the ones of the correct words~\cite{kaneko2020encoder} by keeping the language model frozen, thus producing poor representation in the encoder for such words. Finetuning BERT in a sequence-to-sequence setting results in catastrophic forgetting~\cite{zhu2020incorporating}. We experimented only with BERT weights that were not finetuned. Kaneko et al.~\cite{kaneko2020encoder} recently proposed a way to integrate BERT for GEC, which will be further explored in follow-up experiments.

Future work includes adapting more the artificially generated dataset to the gold corpus by using more similar sentences for the NAC-sentences and NAC-phrases to benefit more from pretraining. This could be done by including other sources for sentences, especially more informal ones. Moreover, we could include non-complete sentences (i.e., phrases without a verb) to resemble more the NAC-phrases group. Also, given the sensibility of GEC training to the gold corpus error rate and the significant differences in POS categories presented in Table \ref{tab:errant-pos}, adjusting the error rates in ERRANT categories of the artificially generated dataset to resemble the error rates of the gold corpus might prove beneficial. In terms of model design, additional techniques proven beneficial in the GEC task could be included, such as dropout on the entire source embeddings~\cite{junczys2018approaching}.

In addition, two issues generated from language specific adaptations need to be taken into consideration. The first one resides in the widespread use of diacritics in Romanian. Approximately one out of three words in regular Romanian texts contains at least a letter with diacritics~\cite{tufics2008diac+}. Although our corpus contains several mistakes related to diacritics, the best model is still unable to correct diacritics errors. Thus, additional corpora obtained in an unsupervised manner by stripping off diacritics should be used for training a model that corrects diacritics. The second issue is related to the richer morphology of the Romanian language, when compared to English. Methods that specifically consider rich morphology \cite{rozovskaya2019grammar} should be also applied for Romanian. 

\section*{Acknowledgments}
This work was funded by “Semantic Media Analytics – SEMANTIC”, subsidiary contract no. 20176/30.10.2019, from the NETIO project ID: P\_40\_270, MySMIS Code: 105976, as well as a grant of the Romanian Ministry of Research and Innovation, CCCDI - UEFISCDI, project number PN-III-P1-1.2-PCCDI-2017-0689 "Revitalizing Libraries and Cultural Heritage through Advanced Technologies", within PNCDI III. This work was performed using TensorFlow Research Cloud programme (\url{https://www.tensorflow.org/tfrc}) which supports computationally-intensive research projects by providing free access to Google's Cloud Tensor Processing Unit (TPU) (\url{https://cloud.google.com/tpu/}).

\bibliography{references}
\bibliographystyle{ieeetr}

\end{document}